\documentclass{article}

\usepackage{microtype}
\usepackage{graphicx}
\usepackage{subfigure}
\usepackage{booktabs}
\usepackage{amsmath} 

\usepackage{hyperref}

\usepackage[accepted]{icml2019}

\begin{document}

\twocolumn[

\title{Domain specificity and data efficiency in typo tolerant spell checkers: the case of search in online marketplaces}
\date{\vspace{-0.2in}}
\maketitle

\icmlsetsymbol{equal}{*}

\begin{icmlauthorlist}
\icmlauthor{Dayananda Ubrangala}{MS}
\icmlauthor{Juhi Sharma}{MS}
\icmlauthor{Ravi Prasad Kondapalli}{MS}
\icmlauthor{Kiran R}{MS}
\icmlauthor{Amit Agarwala}{MS}
\icmlauthor{Laurent Bou\'e}{MS}
\end{icmlauthorlist}

\icmlaffiliation{MS}{Microsoft, CX Data}

\vspace{0.4in}

\begin{abstract}

\vspace{0.4in}

Typographical errors are a major source of frustration for visitors of online marketplaces.  Because of the domain-specific nature of these marketplaces and the very short queries users tend to search for, traditional spell cheking solutions do not perform well in correcting typos.  We present a data augmentation method to address the lack of annotated typo data and train a recurrent neural network to learn context-limited domain-specific embeddings.  Those embeddings are deployed in a real-time inferencing API for the Microsoft AppSource marketplace to find the closest match between a misspelled user query and the available product names.  Our data efficient solution shows that controlled high quality synthetic data may be a powerful tool especially considering the current climate of large language models which rely on prohibitively huge and often uncontrolled datasets.

\vspace{0.4cm}

\end{abstract}
\vspace{0.4in}
]

\icmlcorrespondingauthor{Laurent Bou\'e}{laurent.boue@microsoft.com}

\printAffiliationsAndNotice{}

\section{Introduction}

One of the most common problems that users face while searching for information is typos. Typos, or typing errors, can lead to inaccurate search results and create frustration for the users. Even though search engines use complex algorithms to match the user's search terms with relevant web pages, even minor spelling errors can completely alter the search results.  As such, the question of typo tolerance in search has become a major concern for both users and search engine providers alike. 

We focus on situations where user queries are very domain-specific and tend to be rather short.  This is a common scenario for online marketplaces where users typically search by typing in directly the name of the product they are looking for instead of a grammatically well-formed sentence.

We present a method to identify context-limited typos in domain-specific settings.  Our solution can be split into three parts. First, we analyze and classify real-world typographical errors made by users on other platforms.  These foundational statistics are used to generate synthetic training datasets that are specific to our target corpus of AppSource marketplace product names. Second, we use these datasets to train a multi-layer LSTM model.  Using this trained model, we gather embeddings for the entire AppSource product catalog. Third, those corpus-wide embeddings are compared in real-time with the embeddings of the search query input by the user to get the closest match from product corpora.  

Considering the lack of annotated typo data, our model is trained entirely on synthetically generated datasets. Through progressively more realistic versions of data augmentation strategies, our final model improves the CTR (clickthrough rate) of search results by more than 4\% and decreases the rate of no search results by 8\%. This lift in performance is remarkable in that the model is trained on synthetic data only. 

Our model has been deployed as a real-time API consumed by the Microsoft AppSource marketplace website.  AppSource (formerly known as Office Store) is Microsoft's official marketplace for business applications, add-ins, and content packs that extend the functionality of Microsoft products such as Microsoft 365, Dynamics 365, Power BI, Azure, and more. It provides a platform for developers and partners to publish and distribute their solutions to a wide range of Microsoft customers.  Using AppSource, users can discover and acquire applications and add-ins to enhance their Microsoft productivity and business solutions. These applications range from industry-specific solutions to productivity tools, analytics dashboards, project management tools, customer relationship management (CRM) systems, and more.  AppSource offers a curated collection of trusted applications that have undergone a review process by Microsoft to ensure quality, security, and compatibility. Users can explore various categories, search for specific solutions, read detailed descriptions and reviews, and even try out free trial versions of the applications before making a purchase.  By leveraging AppSource, businesses can extend the capabilities of Microsoft products and tailor them to their specific needs, enhancing productivity and enabling digital transformation within their organizations.  For partners, AppSource represents a highly visible opportunity to showcase and sell their software solutions to a vast customer base, benefiting from the extensive reach and recognition of Microsoft's brand and ecosystem.  Overall, App Source has~$\approx 23,000$ apps in its catalog and the catalog grows roughly by~$\approx 200$ apps per month.

With the deployment of our typo-tolerant spell checker, Microsoft AppSource constitutes one of the few places where non-dictionary-based typo correction systems have been deployed in a production system.

\section{Related work}

Usual spell checkers, such as those found in word processing software, rely on a dictionary-based approach to correct spelling errors.  These lexicons or unigram language models compare the user's input against a pre-existing dictionary of correctly spelled words and flag words that do not match their built-in dictionaries~\cite{hunspell, GNUphonetic}.  Other rule-based systems use word mismatches as defined by traditional natural language processing techniques~\cite{survey, damerau, symmetry, unix, church} to flag potential typos.  Although these approaches are well founded, they are not effective in domain-specific settings because the dictionaries used by common spell checkers are limited to commonly used words and may not include technical terms or jargon specific to a particular field~\cite{diacritics, howDifficult}.  Typically, more modern machine learning based solutions rely on context~\cite{NeuSpell, bert} to identify typos and therefore do not lend themselves to situations of online marketplaces queries which are very short and without much surrounding context.  In fact, we explicitly evaluate the performance of common spellcheckers as exemplified by techniques established in~\cite{pyspellchecker, Norvig}  in Section~\ref{sec:modelArchitecture} and quantify their poor performance.

As noted in~\cite{singleLetter} more than 80\% of errors differ from the correct word by only a single letter.  Furthermore, errors can accurately be classified into just a small number of independent categories~\cite{errorTypes} and several efforts have been made towards generating datasets based on artificial grammatical mistakes~\cite{grammar1, grammar2, grammar3, Ghosh}.  However, real-world typos do not necessarily follow those grammatical constructs~\cite{deMelo, google} and there arises a need to generate synthetic training datasets based on historical typo statistics from open source datasets~\cite{githubData, twitterData}.  This will be the topic of sections~\ref{sec:errorTypeSec} and~\ref{sec:Statistics} of this paper.  Regarding the model architecture, our work most resembles~\cite{rnn} in the use of recurrent neural networks although we use the network in training mode primarily self-supervised~\cite{selfSupervised} on synthetic data to learn domain-specific embeddings~\cite{EmbedJoin}, see Section~\ref{sec:modelArchitecture}.

\section{Classification of typographical errors}
\label{sec:errorTypeSec}

Although one usually refers to typing mistakes under the umbrella term ``typos'', it turns out that typographical errors actually may come under many different guises. Following previous classification studies~\cite{diacritics, singleLetter, errorTypes, deMelo}, we consider the set~$\mathcal{T}$ comprising of one-character typos
\begin{equation}
\mathcal{T} = \big\{ \mathcal{T}_\text{Deletion}, \mathcal{T}_\text{Insertion},\mathcal{T}_\text{Replication} ,\mathcal{T}_\text{Substitution} , \mathcal{T}_\text{Transposition} \bigr\}
\label{eq:errorTypes}
\end{equation}
Using the ground-truth string $s_\text{gt} = \texttt{finally}$ as an example, these~$|\mathcal{T}| = 5$ error types can be illustrated as follows: 
\begin{itemize}
    \item $ \mathcal{T}_\text{Deletion} = \texttt{finlly} \, \neq \, s_\text{gt} $  ; missing one character.
    \item $ \mathcal{T}_\text{Insertion} = \texttt{fainally} \, \neq \, s_\text{gt}$ ; additional character.
    \item $ \mathcal{T}_\text{Replication} = \texttt{fiinally} \, \neq \, s_\text{gt} $ ; special case of~$ \mathcal{T}_\text{Insertion}$ when the added character is the same as its preceding character.
    \item $ \mathcal{T}_\text{Substitution} = \texttt{finelly} \, \neq \, s_\text{gt} $  ; character replaced by another one.
    \item $ \mathcal{T}_\text{Transposition} = \texttt{fianlly} \, \neq \, s_\text{gt} $  ; special case of~$ \mathcal{T}_\text{Substitution}$ when the substituted character takes the place of its neighbor.
    \end{itemize}


\begin{figure}[ht]
\begin{center}
\centerline{\includegraphics[width=\columnwidth]{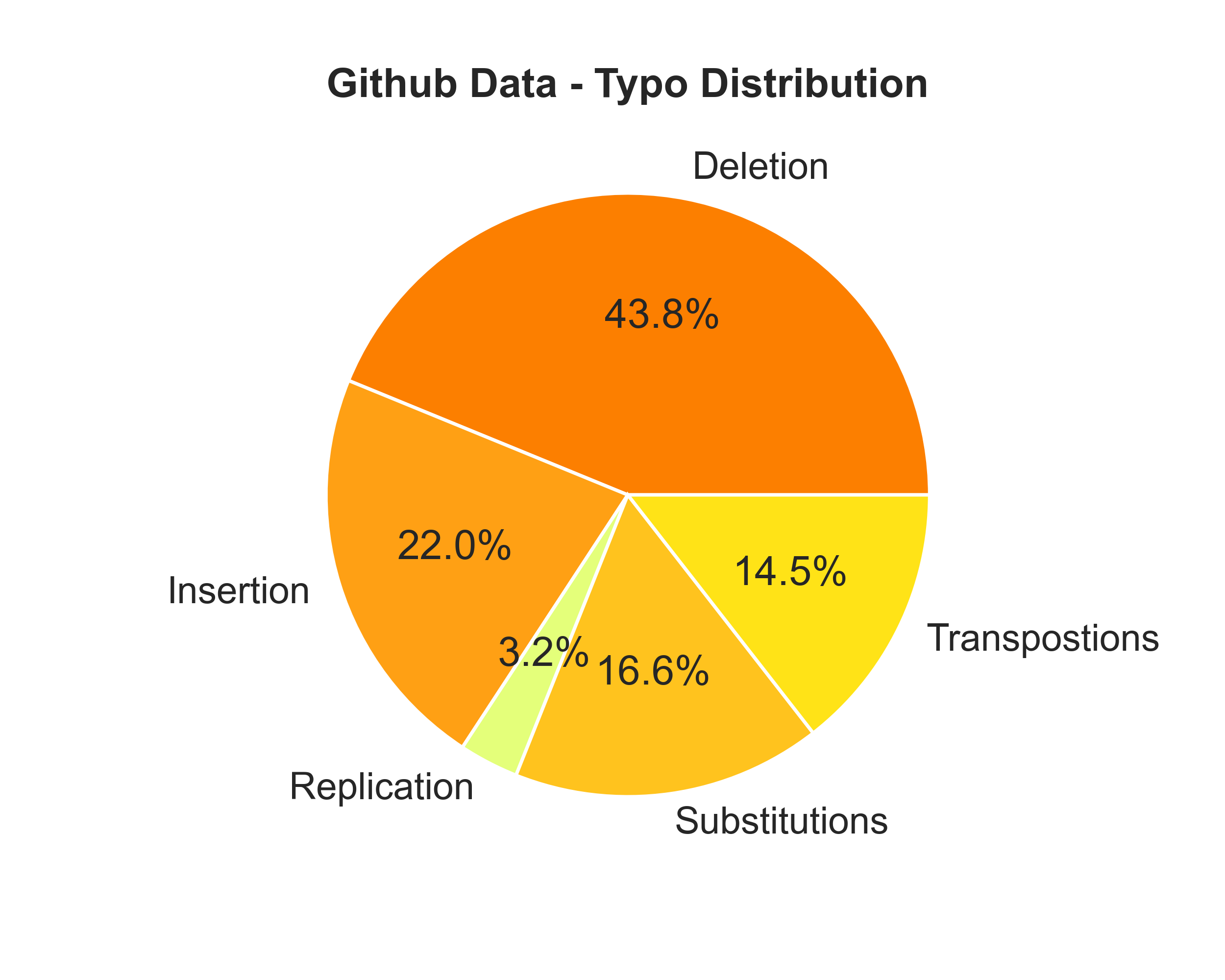}}
\caption{Pie chart representation of the distribution of the different error types~$p_\mathcal{T}\langle \delta = \text{GitHub} \rangle$.  One can see that deletions are far more prevalent than other types of typing errors.}
\label{fig:GitHubDistrib}
\end{center}
\end{figure}

We are considering multiple datasets
\begin{equation}
\delta \in \mathcal{D} = \{ \text{GitHub}~\cite{githubData} , \text{Twitter}~\cite{twitterData} , \text{Proprietary} , \cdots \}
\label{eq:dataSets}
\end{equation}
The Twitter Typo Corpus contains~$\approx 40,000$~pairs of words with typographical errors along with the correct word representing a good variety of typographical errors commonly found in informal social media text.  With about~$\approx 350,000$~edits collected from code commits, the GitHub Typo Corpus is the largest available public typo dataset.  Both of these datasets, along with other proprietary ones we gathered ourselves based on the AppSource search telemetry, are used to gather historical statistical properties that are used in our data augmentation strategies as explained in detail in the following sections.

The distribution of classes of typographical errors for a dataset~$\delta$ is denoted by the~$\mathcal{T}$-dimensional vector
\begin{equation}
p_\mathcal{T}\langle \delta \rangle = \bigg[ p_{\mathcal{T}_i} \langle \delta \rangle \, \vert \, \mathcal{T}_i \in \mathcal{T} \,\,\, \text{and} \,\,\, \sum_{\mathcal{T}_i} p_{\mathcal{T}_i} \langle \delta \rangle = 1  \bigg]
\label{eq:ErrorTypeDistrib}
\end{equation}
where~$p_{\mathcal{T}_i} \langle \delta \rangle$ refers to the probability of observing error type~$\mathcal{T}_i$ in a dataset~$\delta$.  An example of the distribution of these classes of errors can be seen in Fig~\ref{fig:GitHubDistrib} for the GitHub dataset.  

The next step consists in classifying each typo in all datasets~$\mathcal{D}$ into a specific instance~$\mathcal{T}_i \in \mathcal{T}$.  This is achieved by identifying the necessary edits to transform one string (potentially affected by a typo) with another (ground-truth).  This can be done efficiently using standard dynamic programming techniques for sequence matching~\cite{difflib}.

\section{Statistics of typographical errors}
\label{sec:Statistics}

\subsection{Non-locality of the errors}

All instances~$\mathcal{T}_i \in \mathcal{T}$ of error types may emerge from different underlying mechanisms and, as a result, may be characterized by different statistical properties. 

Let us denote by~$\mathcal{K}$ the set of all keys on a keyboard. We define a function~$\mathcal{T}_w$ that takes as argument a dataset~$\delta \in \mathcal{D}$, a class of typo~$\mathcal{T}_i \in \mathcal{T}$ and a keyboard key~$\kappa  \in\mathcal{K}$ and returns a dependently-typed object $\mathcal{P}_w\langle \delta, \mathcal{T}_i , \kappa \rangle$ such that
\begin{equation}
\mathcal{T}_w  : \bigg[ \big( \delta \in \mathcal{D} \big) \times \big( \mathcal{T}_i \in \mathcal{T} \big) \times \big( \kappa  \in\mathcal{K} \big) \bigg] \rightarrow \mathcal{P}_w \langle \delta, \mathcal{T}_i , \kappa \rangle
\label{eq:errorStats}
\end{equation}
where $\mathcal{P}_w \langle \delta, \mathcal{T}_i, \kappa \rangle$ may be either:
\begin{itemize}
    \item $\mathcal{P}_w\langle \delta,\mathcal{T}_i , \kappa\rangle \in [0,1]$ if~$\mathcal{T}_i \in \{ \mathcal{T}_\text{Deletion}, \mathcal{T}_\text{Insertion} \}$. In this case $\mathcal{P}\langle \delta,\mathcal{T}_i , \kappa\rangle$ is a constant that encodes the probability of deletion / insertion of the key~$\kappa$.
    \item $\mathcal{P}_w\langle \delta, \mathcal{T}_i , \kappa\rangle = \big[ p_1, \cdots , p_{\mathcal{|\mathcal{K}|}} \big] \in \big[ [0,1] \times \cdots \times [0,1] \big]$ if $\mathcal{T}_i \in \{ \mathcal{T}_\text{Replication}, \mathcal{T}_\text{Substitution}, \mathcal{T}_\text{Transposition} \}$.  In this case $\mathcal{P}\langle \delta,\mathcal{T}_i , \kappa \rangle$ is a probability density function such that~$\sum_{\kappa^\prime \in \mathcal{K}} p_{\kappa^\prime} = 1$.  It represents the probability of replicating / substituting / transposing the initial key~$\kappa$ by any other key~$\kappa^\prime \in \mathcal{K}$.
\end{itemize}
In practice the function~$\mathcal{T}_w$ is implemented efficiently using nested key-value data stores.  

\begin{figure}[ht]
\begin{center}
\centerline{\includegraphics[width=\columnwidth]{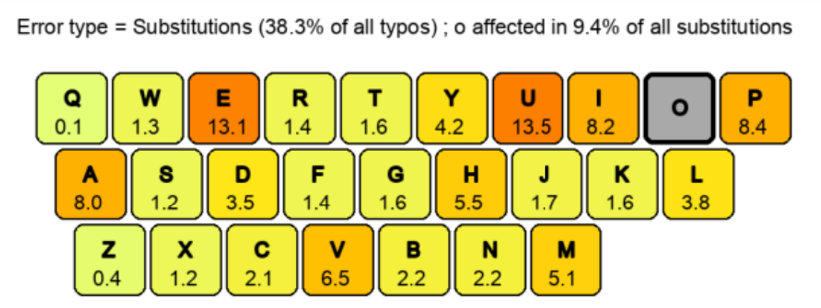}}
\caption{Heatmap representation of the probability distribution obtained by evaluating~$\mathcal{T}_w (\text{GitHub}, \mathcal{T}_\text{Substitution}, \text{``o''} ) $.  Colors represent the probabilities~$p_{\kappa^\prime}$ of substituting~$\kappa = \text{``o''}$ by any other other key~$\kappa^\prime \in \mathcal{K}$. }
\label{fig:exampleTypos}
\end{center}
\end{figure}

We populate our~$\mathcal{P}_w \langle \delta,\mathcal{T}_i , \kappa \rangle$ statistics completely on real-world examples of typos. For illustration purposes, we show in~Fig~\ref{fig:exampleTypos} a typical probability density function of keystroke mistakes estimated from the GitHub dataset.  The non-local effects are clearly visible with many keys~$\kappa^\prime$ physically far away from~$\kappa$ being attributed higher probabilities of substitutions than those closer to it.  Among other causes, this may happen due to language/phonetic effects such as ``farward'' instead of ``forward'' or ``thaought'' instead of ``thought''...

At any rate, this observation invalidates the assumptions of keyboard locality implied in the QWERTY distance (and its derivatives) showing that non-local effects are very strong and should not be ignored.  Taking these into account thanks to our sophisticated~$\mathcal{T}_w$ is what allows us to create a more powerful synthetic data augmentation strategy.

\subsection{Position distribution of the errors}

The character position~$r$ at which the typographical errors occur is another random variable characterizing the statistics.  We normalize~$r$ by the length of the mistyped string so that strings of any lengths can be compared to each other ($r=0$ always corresponds to the first character and~$r=1$ coincides with the last character).

Obviously, the statistics of~$r$ may depend on the class of typo and following the notation from the previous section, we denote by~$\mathcal{T}_r$ the function  
\begin{equation}
\mathcal{T}_r  : \bigg[ \big( \delta \in \mathcal{D} \big) \times \big( \mathcal{T}_i \in \mathcal{T} \big) \big) \bigg] \rightarrow \mathcal{P}_r \langle \delta , \mathcal{T}_i \rangle
\label{eq:posStats}
\end{equation}
where~$\mathcal{P}_r \langle \delta , \mathcal{T}_i \rangle$ is a probability distribution that quantifies the likelihood of relative character position~$r$ being affected by an error of type~$\mathcal{T}_i$ for dataset~$\delta$.

\begin{figure}[ht]
\begin{center}
\centerline{\includegraphics[width=\columnwidth]{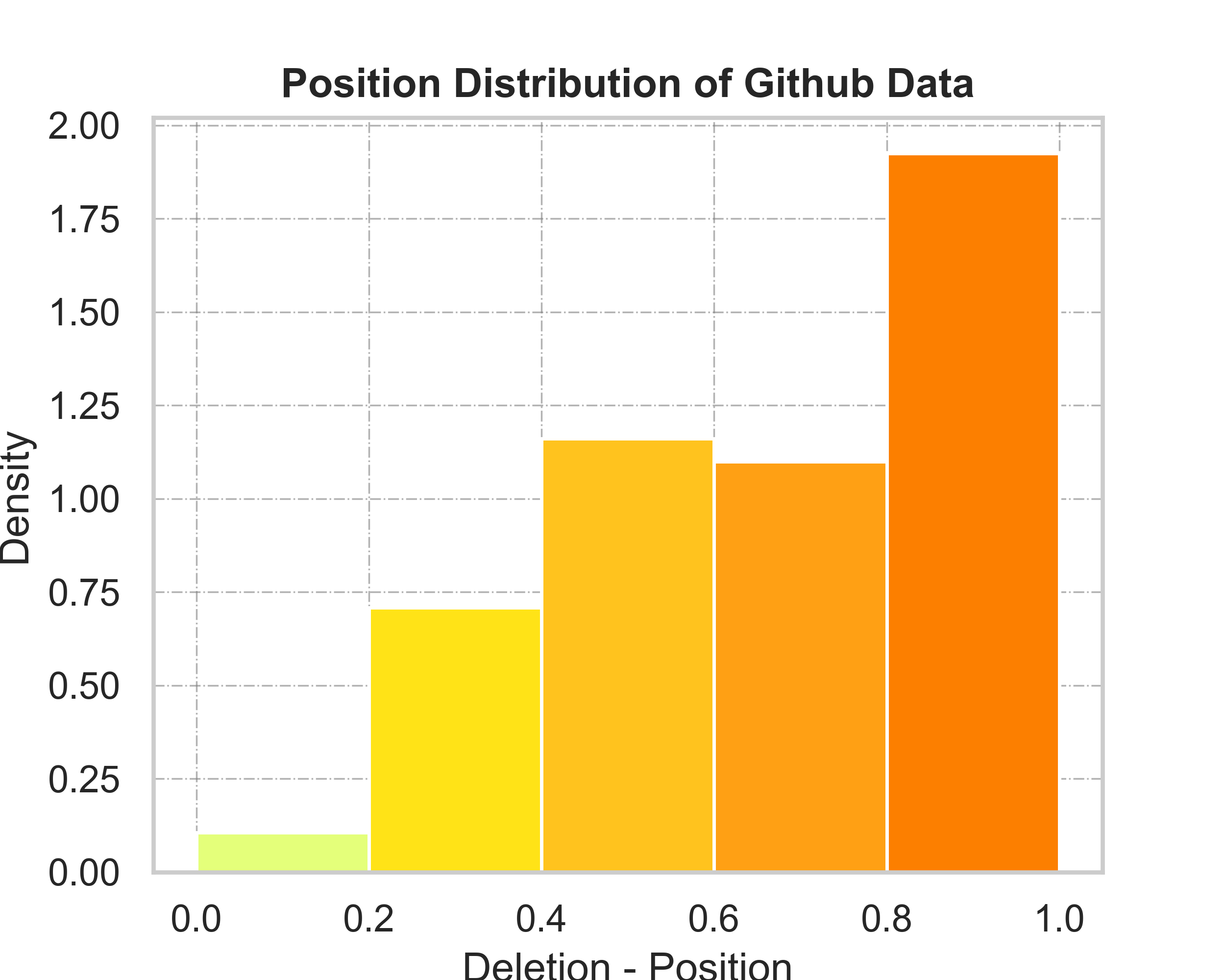}}
\caption{Illustration of non-uniform effects in keystroke position mistake distribution.  As an example, we see that the probability of key deletion increases linearly as a function of letter position:  letters are far more likely to be deleted towards the end of words than they are at the beginning.}
\label{fig:githubPosDistb}
\end{center}
\end{figure}

As we can see~in~Fig~\ref{fig:githubPosDistb} for deletions,~$\mathcal{P}_r \langle \delta , \mathcal{T}_i \rangle$ does not follow a uniform distribution.  The same observation carries over for the other classes of typos~$\in \mathcal{T}$ as well and those statistical properties will be taken into account in our synthetic datasets.

\section{Typo correction ML formulation}
\label{sec:modelArchitecture}

Before we move on to the different data augmentation strategies and their relative performance, we briefly describe the formulation of our typo correction solution.

Common spellcheckers which are typically built on top of Levenshtein-like distances such as the ones used in many Microsoft products are not accurate enough for the short and domain-specific queries specialized online marketplaces such as AppSource face even if their dictionaries are regularly updated.  As an example, we have used the popular open-source package pyspellchecker which works by comparing permutations within a predefined Levenshtein distance.  When trained only on default dictionaries, the spell checker achieves only a very small accuracy of~$18.3\%$.  Even when the dictionary is enhanced with product names from the AppSource catalog, the accuracy reaches only~$59.9\%$ which is well below our baseline model (see Table~\ref{tab:perf}).  Considering the poor performance of traditional spellcheckers, we now introduce our formulation of domain-specific typo correction as a multi-class classification problem.

\subsection{Training: multiclass classification}
\label{sec:TrainingModel}

We start by training a supervised classification model with~$|\mathcal{V}| \approx 23,000$ classes corresponding to the product names in the AppSource marketplace catalog.  The details of model architecture are shown in Fig.~\ref{fig:modelArchitecture}. As the focus of the present study is about characterizing different types of data augmentation strategies and their performance, we limit ourselves to relatively small and simple recurrent networks upon which we can iterate quickly.  Once this model has been trained, we use it as a proxy from which we can extract the domain-specific ``embedding''~\footnote{By ``embedding'', we refer to the feature map at the last layer before the softmax activation as is common terminology in the literature.} representations for the~$|\mathcal{V}|$ product names which we cache into a database.

\subsection{Inference: nearest neighbor in embedding space}
\label{sec:Inference}

When users type in a query, the embedding representation of this query can be compared to our database embeddings of~$\mathcal{V}$ and the nearest neighor (as measured by cosine similarity) is returned as the ``predicted'' class.  In the special case where the user query matches exactly an existing product name, the similarity will be exactly~1 as expected and this similarity score will then decrease as typos get more and more different from the product names in~$\mathcal{V}$.

\subsection{Model performance evaluation}
\label{sec:ModelEval}

Using historical production web telemetry data, we extracted~$3,303$ of the most common user queries which we identified as being typos with respect to an existing product name in the AppSource marketplace catalog.  Then, we manually labeled each one of these typos with the correct product that the user eventually clicked on.  This process enabled us to build a validation dataset upon which the accuracy of the model can be evaluated in the inference mode described in section~\ref{sec:Inference}.  Accuracy is simply defined as the number of times the model predicts the correct class normalized by~$|\mathcal{V}|$. 

\begin{figure}[ht]
\begin{center}
\centerline{\includegraphics[width=\columnwidth]{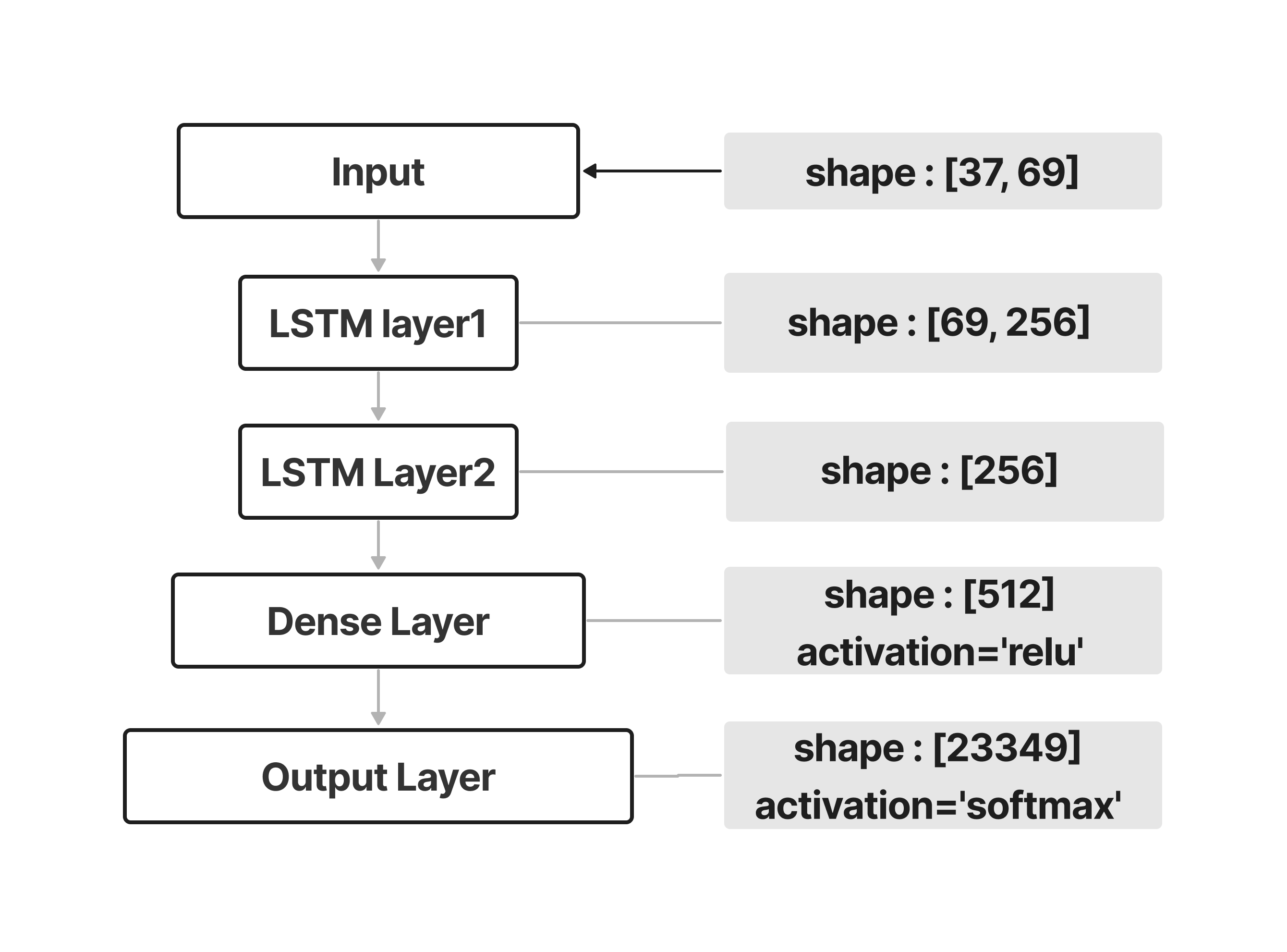}}
\caption{We consider product names as a sequence of characters of length~$s = 69$ (maximum product name length) where each character is represented as a one-hot-encoded vector of size~$f = 37$ (total number of distinct characters). Batching together~$n = 128$ products, the input data of shape~$\sim (n\times s \times f)$ is fed into two LSTM layers which return a recurrent hidden layer of shape~$\sim (n\times s \times h)$ with~$h=256$.  Eventually, this representation is flattened and connected to a dense~512 neuron layer with RELU activation before a final dense layer with softmax activation for classification into~$|\mathcal{V}| = 23,349$ classes using the cross-entropy loss.  The model was trained with an Adam optimizer with learning rate of 0.001 over 50 epochs.  The ``embedding'' representation of input string is a~512 dimensional vector.}
\label{fig:modelArchitecture}
\end{center}
\end{figure}

\section{Training on completely synthetic data}

\begin{algorithm}[tb]
   \caption{Synthetic training dataset generation}
   \label{alg:synthetic}
\begin{algorithmic}
   \vspace{0.15cm}
   \STATE {\bfseries Inputs} Ground-truth string $s_\text{gt}$, desired number of synthetic samples $N$, classes of typos $\mathcal{T}$, historical statistics $\mathcal{T}_w$ and $\mathcal{T}_r$, $\delta$ dataset
   \vspace{0.15cm}
   \STATE {\bfseries Output:} Synthetic samples $\mathcal{S} = [ s_1 , \cdots, s_N ]$    
   \vspace{0.15cm}
   \STATE Set \texttt{keepDuplicate = True}
   \STATE Initialize $\mathcal{S} \leftarrow [\,]$ and $i \leftarrow 1$
   \vspace{0.1cm}
   \WHILE{$i \leq N$}
   \vspace{0.1cm}
   \STATE Pick random $\mathcal{T}_i$ with probability given by~Eq.~(\ref{eq:errorStats}).
   \STATE Pick random $r$ with probability given by~Eq.~(\ref{eq:posStats}).
   \STATE $\kappa \leftarrow$ character from $s_\text{gt}$ picked from the 2 steps above.
   \STATE Apply appropriate action on~$\kappa$
   \STATE $s_i \leftarrow$ Associate synthetic sample with label~$s_\text{gt}$
   \IF{$s_i \in \mathcal{S}$ and \texttt{keepDuplicate = False}}
   \STATE $i \leftarrow i - 1$ \\ \bfseries{continue}
   \ENDIF
   \STATE Add~$s_i$ to $\mathcal{S}$
   \STATE $i \leftarrow i + 1$
   \ENDWHILE
\end{algorithmic}
\end{algorithm}

\begin{table}[t]
\begin{center}
\begin{small}
\begin{sc}
\begin{tabular}{|l|r|}
\toprule
Training dataset & Accuracy in \% \\
\midrule \midrule
Basic spellchecker (\ref{sec:modelArchitecture}) & $18.3$ \\
Specialized spellchecker (\ref{sec:modelArchitecture}) & $59.9$ \\
\midrule \midrule
Random (\ref{sec:pureRandom}) & $62.37$ \\
QWERTY-distance (\ref{sec:QWERTY}) & $62.25$ \\
\midrule \midrule
Real-World Statistics (\ref{sec:RWS}) &  \\
\midrule
$\delta =$ GitHub  & 65.06  \\
$\delta =$ Twitter  &  64.03 \\
$\delta =$ Proprietary  & 64.27 \\
\midrule
$\delta =$ w/o duplicate samples  & 63.88 \\
\midrule
\midrule
$\delta =$ Dataset fusion (\ref{sec:Fusion}) & 65.58  \\
\bottomrule
\end{tabular}
\end{sc}
\end{small}
\end{center}
\caption{Performance comparison for different type of training datasets.  One can see that the model performance significantly improves as one is introducing more and more sophisticated data augmentation strategies.}
\label{tab:perf}
\end{table}

As discussed in the introduction, we are facing an unusual situation where there is no training data other than the ground-truth vocabulary~$\mathcal{V}$ of product names.  Therefore, if one is to train the supervised machine learning model specified in section~\ref{sec:TrainingModel}, we have to resort entirely on creating a synthetic training dataset.

We consider multiple stages of sophistication in creating such synthetic data and demonstrate via careful experiments that the model performance can be significantly improved by gradually introducing more realistic synthetic data.

All augmentation strategies presented below follow the same procedure of generating a dataset by running algorithm~(\ref{alg:synthetic}) on all product names from~$\mathcal{V}$.  This creates a list of~$n \times |\mathcal{V}|$ samples where each class (i.e. product name) has~$n$ (potentially duplicated; see below) synthetic samples associated with it.  Eventually, this synthetic dataset is used to train the supervised model of~Fig.~\ref{fig:modelArchitecture}.  The accuracy of the model is estimated on the manually annotated dataset described in Section~\ref{sec:ModelEval}.

Because those synthetic datasets are created directly from product names in~$\mathcal{V}$, they are, by construction,  domain-specific to this catalog.

\subsection{Random augmentation}
\label{sec:pureRandom}

The first stage consists taking a uniform distribution over the error types~$\mathcal{T}$ defined in~Eq.(\ref{eq:errorTypes}) and forcing their respective statistics~$ \mathcal{P}_w \langle \delta, \mathcal{T}_i , \kappa \rangle$ and~$ \mathcal{P}_r \langle \delta, \mathcal{T}_i \rangle$ to be simple uniform random variables.  With this statistical set-up in place, we follow algorithm~(\ref{alg:synthetic}) to generate a synthetic dataset and train the model. This augmentation strategy leads to a performance~$62.37\%$ (see table~\ref{tab:perf}).

\subsection{QWERTY-distance based augmentation}
\label{sec:QWERTY}

Here, once again the error types are drawn from a uniform distribution over~$\mathcal{T}$.  The only difference from the purely random augmentation of section~\ref{sec:pureRandom} is that the pair of keys involved in substitutions are now limited to nearby keys on the physical keyboard.  In practice, given a key~$\kappa$, we limit the possible substitutions to keys that are a QWERTY distance of one compared to~$\kappa$.   This is akin to a weighted Levenshtein distance where only the keys immediately surrounding the key~$\kappa$ of interest are assigned equal and non-zero weights.  This augmentation strategy leads to a performance~$62.25\%$ (see table~\ref{tab:perf}).

\subsection{Real-world statistics}
\label{sec:RWS}

In this case, we use the real-world distribution of error types~$p_\mathcal{T}\langle \delta \rangle$ discussed in section~\ref{sec:errorTypeSec} along with their appropriate observed historical statistics~$ \mathcal{P}_w \langle \delta, \mathcal{T}_i , \kappa \rangle$ and~$ \mathcal{P}_r \langle \delta, \mathcal{T}_i \rangle$ described in section~\ref{sec:Statistics} to generate the synthetic training dataset.  

The model performance is significantly improved for all 3 independent datasets in~$\delta \in \mathcal{D}$ as one can see in table~\ref{tab:perf} with the best performance of~$65.06\%$.

Note that we kept open the possibility of removing duplicate synthetic samples in algorithm~(\ref{alg:synthetic}) by controlling the value~\texttt{keepDuplicate}.  Intuitively, one should expect model performance degradation by removing the duplicated samples as their removal would create a bias away from historical statistics.  Indeed this is what we observed with the best performance without duplicates reaching only~$63.88\%$.

\subsection{Hyperparametrized dataset fusion}
\label{sec:Fusion}

\begin{figure}[ht]
\begin{center}
\centerline{\includegraphics[width=\columnwidth]{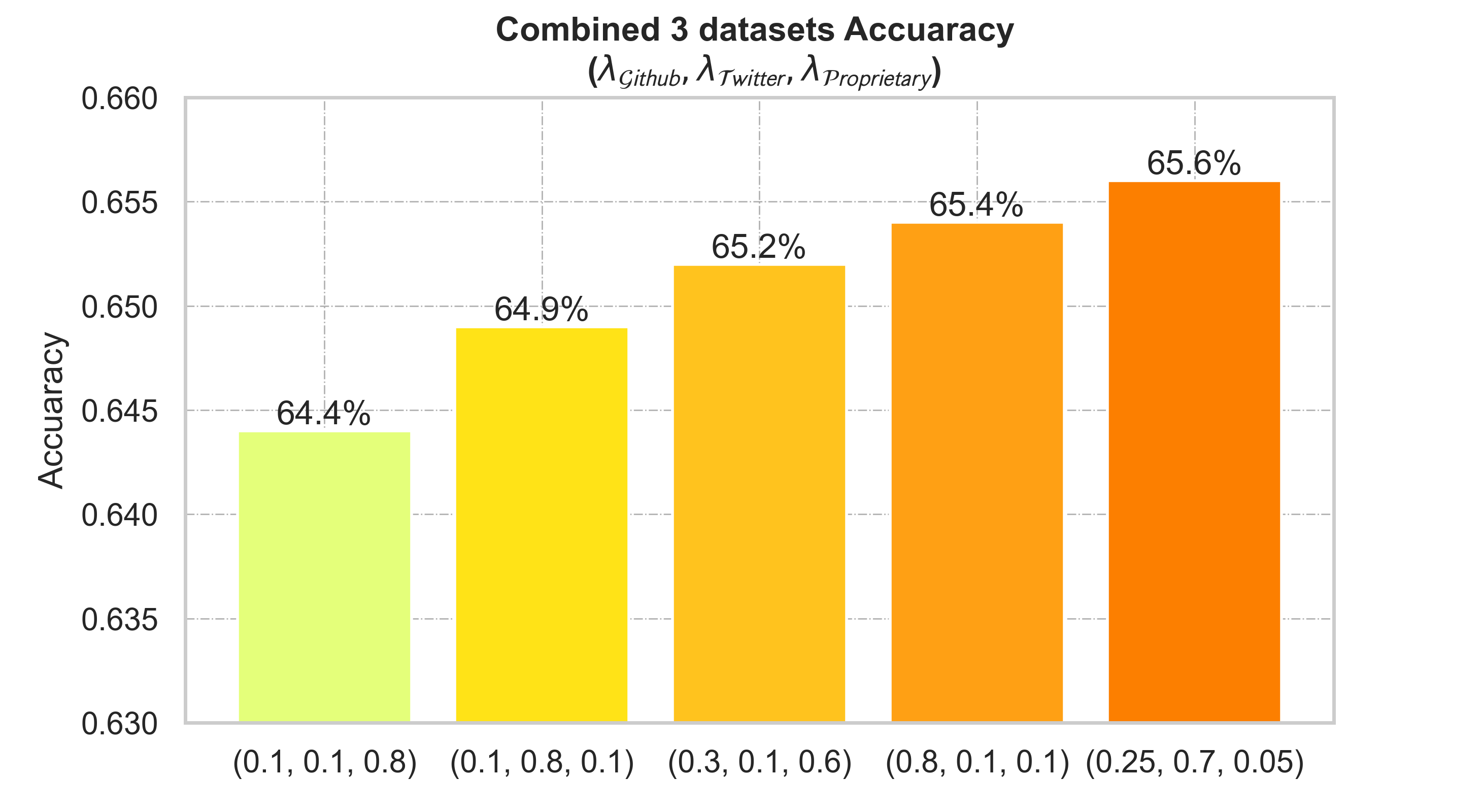}}
\caption{Model performance vs. different mixing ratios of data fusion for~$|\mathcal{D}| =3$ datasets.  One can see that the best accuracy is reached for~$\{ \lambda_\text{GitHub} = 0.25, \lambda_\text{Twitter} = 0.7, \lambda_\text{Proprietary} = 0.05\}$.}
\label{fig:dataFusion}
\end{center}
\end{figure}

The previous strategy was based on drawing the typo statistics from a single dataset~$\delta$ at a time.  It may be that some aspects of our unique AppSource marketplace situation are better represented by some datasets than others.  In order to potentially take the best out of all the available datasets, we propose to fuse the statistics of the datasets of~$\mathcal{D}$ together by introducing hyperparameters.

Given a sets of datasets~$\delta \in \mathcal{D}$, see~Eq.(\ref{eq:dataSets}), we combine them by introducing dataset-dependent hyperparameters~$\lambda_\delta$ such that the final combined dataset is a linear mixture
\begin{equation}
\mathcal{D}_\text{dataset fusion} = \big\{ \lambda_\delta \times \delta \,\, \big| \,\,  \sum_{\delta \in \mathcal{D}} \lambda_\delta = 1 \big\}
\end{equation}
Using grid search for hyperparameter tuning, we observe that this optimization is indeed successful in creating more appropriate training data leading to an eventual model accuracy of~$65.58\%$ (see Fig.~\ref{fig:dataFusion} and table~\ref{tab:perf}).

\subsection{Data efficiency}

\begin{figure}[ht]
\begin{center}
\centerline{\includegraphics[width=\columnwidth]{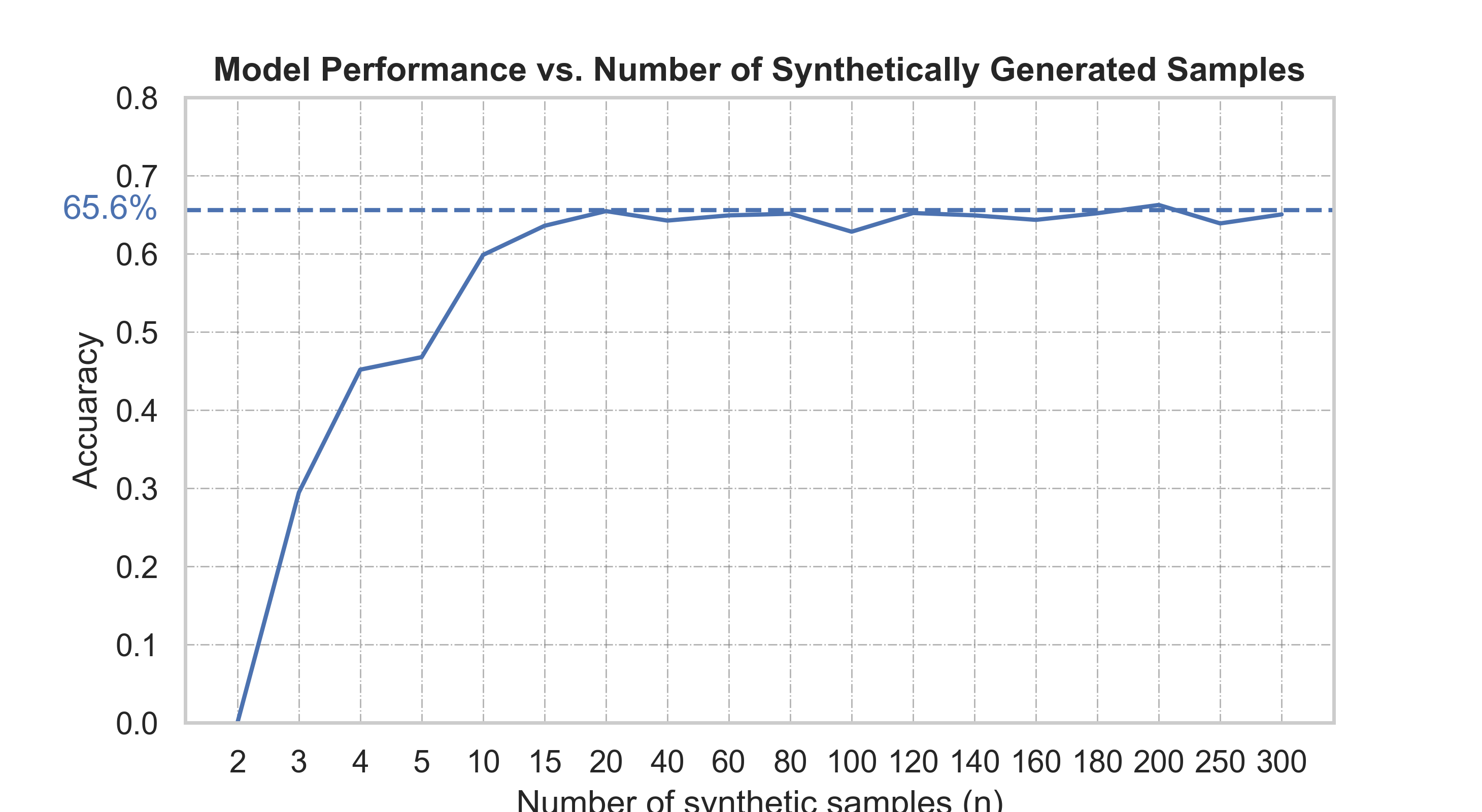}}
\caption{Model performance vs. number of synthetically generated samples showing excellent convergence properties with minimal number of synthetic samples.}
\label{fig:sample_size_tune}
\end{center}
\end{figure}

Finally, we conclude this section by commenting on the data efficiency of our augmentation strategy.  It turns out that model performance already saturates and reaches its maximum plateau after only about~$\approx$~20 synthetic samples as demonstrated in~Fig.~\ref{fig:sample_size_tune}.  This quick convergence rate can be related to the average number of characters~$\approx 24$ of the product names in the AppSource marketplace catalog.  

\section{Model deployment as a real-time API}

The model is exposed to the AppSource marketplace team via a real-time API which receives~$\approx 100,000$ daily requests.  Model inference takes around~$\approx 400$ milliseconds along with another 100 milliseconds for API call (including load balancing and traffic management). This means that the total response time is around~$\approx 500$ milliseconds. Based on telemetry logs since Feb. 2023, we have seen that~$99.9\%$ of the API calls to our real-time API are getting a response below 500 millseconds, which is meeting the SLA (Service Level Agreement) with our downstream stakeholders.  Since the AppSource volume is now~$\approx 23,000$ and growing at~$\approx 200$ apps/month, we estimate that the current solution would continue to meet SLAs for at least another 2 years of projected volume of apps.  In the future, we intend to explore vector databases and faster similarity search techniques to handle the performance for even larger values of~$|V|$.

Even though the primary search engine powering AppSource is Azure Cognitive Search (ACS)~\cite{ACS}, this solution frequently fails to return any results and/or any auto-completion for uncommon search queries.  When this happens, our API is triggered and returns the closest matched keyword from the catalog~$\mathcal{V}$. This keyword is further passed back to~ACS thereby providing incremental benefit on top of the default search engine. After deployment of our model, the~CTR (Click Through Rate) improved by~4\% (from 35\% to 39\%) and no-results searches dropped by~8\% (from 25\% to 17\%). Azure Traffic Manager is leveraged for load balancing the search requests across 4 regions: US, Europe, Japan and Australia.

\section{Conclusion}

Solving typos in search is a complex task, particularly in domain-specific settings, because the search terms used in these settings can be highly specialized and technical in nature. Domain-specific search terms are often used by professionals in their respective fields and may include scientific terms, jargon, or acronyms that are not commonly used in everyday language.

We have introduced a domain-specific typo correction model which is completely based on synthetic training data.  We have shown that gradually introducing more sophisticated data augmentation strategies led to significantly better model accuracy.  We have also demonstrated that the data augmentation strategy is very efficient in terms of data size.

The model has been deployed as a real-time API now powering the AppSource marketplace website which is a major portal for customers as well as Microsoft partners.  On average, 50 products get added every week to the AppSource Product catalog and hence our model is trained every week to learn about the newly added products.  The model has already been attributed to significant search improvements in monitored metrics such as~CTR and 0-search results.

In the future, we intend to expand our universe of error types to include multiple-letter typos and incorporate phonetic language effects into our synthetic data augmentation scheme~\cite{GNUphonetic, phonetic2}.  Continuing to gradually increase levels of sophistication of data augmentation, one could consider fully hyperparametrized statistics that are no longer drawn from historical datasets.  In this case, the requirements that error type distributions form a well defined probability distribution could even be lifted leading to more flexibility and potentially higher model accuracy.  Now that we have established the usefulness of our data augmentation strategy to improve the spell-cheking performance, we intend to experiment with more sophisticated network architectures (such as transformer-based) that go beyond our initial recurrent networks.

More generally, our work demonstrates that completely synthetic datasets can be successful in real-world applications that require high levels of accuracy.  Longer term, we hope that more frequent use of synthetic data will enable rapid experimentation and testing as well as reduce the risks of privacy violations associated with using sensitive real-world data. 

\section{Acknowledgements}

We thank our colleagues in CX Data for their feedback and support. In particular, we thank Manish Shukla, Daniel Yehdego, Yasaswi Akkaraju and Naveen Panwar for initiating earlier versions of this model.  Additionally, we thank Noam Ferrara, Gal Horowitz and Greg Oks from the marketplace engineering team for integrating our real-time API into the overall search flow architecture.

\bibliographystyle{ieeetr}
\bibliography{typo}

\end{document}